\documentclass[journal,draftcls,onecolumn,12pt,twoside]{IEEEtran}

\usepackage{hyperref}       
\usepackage{url}            
\usepackage{booktabs}       
\usepackage{amsfonts}       
\usepackage{nicefrac}       
\usepackage{microtype}      
\usepackage{float} 
\usepackage{lipsum}

\usepackage{algpseudocode}
\usepackage{algorithm}

\usepackage{epsfig, graphicx, amsmath, amssymb,cite}
\usepackage{amsthm}
\usepackage{lipsum}
\usepackage{mwe}

\DeclareMathOperator*{\argmin}{\arg\!\min}

\IEEEoverridecommandlockouts
\usepackage[font=small,skip=5pt]{caption}
\begin{document}
\title{\huge Learning End-to-End Codes for the BPSK-constrained Gaussian Wiretap Channel}

\author{Alireza Nooraiepour and Sina Rezaei Aghdam\thanks{A. Nooraiepour
is with WINLAB, Department of Electrical and Computer Engineering,
Rutgers University, NJ, USA. S. Rezaei Aghdam is with the Department of Electrical
Engineering, Chalmers University of Technology, Gothenburg, Sweden (e-mails: alireza.nooraiepour@rutgers.edu, sinar@chalmers.se).}}
\date{}
\maketitle

\begin{abstract}
Finite-length codes are learned for the Gaussian wiretap channel in an end-to-end manner assuming that the communication parties are equipped with deep neural networks (DNNs), and communicate through binary phase-shift keying (BPSK) modulation scheme. The goal is to find codes via DNNs which allow a pair of transmitter and receiver to communicate reliably and securely in the presence of an adversary aiming at decoding the secret messages. Following the information-theoretic secrecy principles, the security is evaluated in terms of mutual information utilizing a deep learning tool called MINE (mutual information neural estimation). System performance is evaluated for different DNN architectures, designed based on the existing secure coding schemes, at the transmitter. Numerical results demonstrate that the legitimate parties can indeed establish a secure transmission in this setting as the learned codes achieve points on almost the boundary of the equivocation region.  
\end{abstract}

\section{Introduction}\label{Sec:Introduction}
Physical layer (PHY) security has been put forth as an alternative/aid to the higher-layer security approaches including cryptography in order to relieve the burden placed by them upon the communication systems in various ways. Wiretap channel \cite{OurSurvey} is a widely-known theoretical model for studying PHY security from an information-theoretic perspective. The importance of achieving physical layer security for this model through finite alphabet signaling like binary phase-shift keying (BPSK) modulation is highlighted in many works \cite{OurSurvey,lowerbound2}. Several works have studied this channel and designed coding schemes to ensure security for that. Specifically, the authors in \cite{nonsys} have proposed an encoding technique called scrambling which could result in BERs very close to $0.5$ for the eavesdropper (Eve) through error propagation, while ensuring a reliable communication for Alice and Bob known as the legitimate parties. The coset-coding approach has been studied for the Gaussian wiretap channel under the name of the randomized scheme in several works \cite{OurSurvey,myTCOM}. The authors in \cite{myTCOM,RandomizedTurbo} utilize convolutional and turbo codes for the randomized scheme and propose bounds on the performance of the optimal decoders in this setup. Furthermore, the application of low-density parity-check (LDPC) codes to the randomized scheme is studied in \cite{RSCLDGM}. The metric used to evaluate the coding performance in these works is security gap which reflects the required difference between the qualities of the Bob's and Eve's channels in order to comply with the security and reliability criteria both measured in BER. The major shortcoming of this metric is that it assumes a BER of $0.5$ at Eve is an indication of a secured system, however, it does not provide any insight to the information-theoretic nature of the wiretap channel. In an effort to make such a connection, the authors in \cite{lowerbound1} obtain a lower bound on the equivocation rate, i.e., the conditional entropy of the secret messages given the Eve's observation, 
and obtain achievable points on the equivocation region for several LDPC codes. 

Application of deep learning for the Gaussian wiretap channel is studied in \cite{Deepshit} where the authors resort to an alternative metric for measuring security claiming that the computation of mutual information is not tractable. The most similar work to the one presented in this paper is \cite{autoencoderPHY} where the authors utilize autoencoders for designing wiretap codes by leveraging a loss function which captures both reliability and security constraint. However, they rely on approximations to compute the mutual information and limit their analysis to the low SNR regimes. Furthermore, their approach does not address the problem of finite alphabet signaling for the wiretap channel.

In this work, we propose learning practical codes for the wiretap channel in an end-to-end manner through stochastic gradient descent (SGD) algorithms using mutual information as the secrecy metric. In this way, we are able to find the true characterization of the system from an information-theoretic perspective. To this end, the encoder at Alice and the decoders at Bob and Eve are represented by deep neural networks (DNNs) which are trained in order to achieve reliable and secure communication for the Alice-Bob link while Eve is trained to decode the secret messages. These goals are further translated to appropriate loss functions which are to be minimized over the DNNs' parameters via SGD algorithms. In order to comply with the BPSK-constrained wiretap channel, we force the DNN at the encoder to produce binary values, and send $+1$ an $-1$ through the channel. The loss function for Alice-Bob includes the BER at Bob and the mutual information between the Eve's observation and the secret messages, which correspond to reliability and security constraints, respectively. Regrading the security loss, we use a recently-proposed deep learning algorithm called mutual information neural estimation (MINE) \cite{MINE} to directly compute the mutual information for our setup. Moreover, as Eve's goal is to decode the secret messages as well, its loss function is chosen to be the corresponding BER at Eve. We investigate the system performance for different DNN architectures at Alice, which are inspired by the existing secure coding schemes. Numerical results illustrate that a DNN which incorporates random bits for encoding the data bits substantially outperforms the other encoders, and the corresponding learned codes achieve points as close as $0.003$ from the boundary of the equivocation region. We also compare our result with the autoencoder approach presented in \cite{autoencoderPHY} and the polar wiretap code \cite{polar} as a state-of-the-art reference points. This comparisons show that our proposed structure for the DNNs in an end-to-end learning setting outperforms the existing methods in terms of reliability and secrecy.

The paper is organized as follows. The system model is presented in Section \ref{sec:Channel Model}. An overview of two of the most effective codes for the wiretap channel is discussed in Section \ref{sec:Coding schemes for securing the PHY layer}. Deep learning techniques required for the end-to-end learning of the codes including MINE is described in Section \ref{sec:Deep learning tools for code design}. We propose end-to-end learning of the codes for the wiretap channel in Section \ref{Sec:Secure Code Design via Deep learning}. Numerical results are presented in Section \ref{sec:Numerical Results}. Finally, the paper is concluded in Section \ref{sec:Conclusions}.  

\section{System Model} \label{sec:Channel Model}
The Gaussian wiretap channel is considered where the channels corresponding to Alice-Bob and Alice-Eve are assumed to be additive white Gaussian noise (AWGN). The former is referred to as the main channel while the latter is called the wiretapper channel. Denoting an input of length $n$ by $\mathbf{x}$, the output of an AWGN channel is obtained by $\mathbf y =\mathbf x+\mathbf w$, where $\mathbf w$ is a length-$n$ Gaussian noise vector whose components are independent and identically distributed (i.i.d.) with variance $N_0/2$ and zero mean. Also, $\mathbf{x}$ denotes the binary phase-shift keying (BPSK) version of the transmitted codeword $\mathbf{c}$. When energy per dimension is one, we have $E_b=1/R$ where $E_b$ and $R$ denote the energy per bit and transmission rate, respectively. We refer to $E_b/N_0$ as signal-to-noise ratio (SNR) in the rest of the paper.
The designer's goal in this model is to come up with codes that satisfy two criteria: reliability and security. The former ensures that the secret messages $\mathbf m$ can be decoded with sufficient reliability at Bob, i.e., BER$_B\leq \kappa$. The latter expresses the requirement that Eve should not be able to extract information about the secret messages. This is captured through $\mathbb{I}(\mathbf m;\mathbf{y}_E)\leq\epsilon$ where $\mathbf{y}_E$ denotes  Eve's observation. The parameters $\kappa$ and $\epsilon$ are predefined small values that are set by the designer to meet the system's requirement. In this work, we aim at learning end-to-end codes in order to satisfy these requirements. To this end, we assume each party is equipped with DNNs for encoding/decoding. The objective for Alice-Bob is to learn codes that satisfy the reliability and security constraints while Eve is trained to decode the secret messages.
\section{Coding Schemes for Securing the PHY Layer}
\label{sec:Coding schemes for securing the PHY layer}
Researchers have relied on tools and ideas from cryptography and coding theory literature to design secure codes for the wiretap channel. In this section, we introduce two of the most effective coding schemes, i.e., scrambling and randomized encoding, which reportedly result in the lowest security gaps \cite{nonsys,myTCOM}. Later on in Section \ref{Sec:Secure Code Design via Deep learning}, we leverage the main ideas behind these techniques in order to design the DNN at Alice. In the classical coding approach, a data message $\mathbf{u}$ of length $k$ is mapped to a codeword $\mathbf{c}$ of length $n$ through $\mathbf{c}=\mathbf{u}\mathbf{G}$, where $\mathbf{G}$ is a $k\times n$ matrix. $\mathbf{G}$ is called the generator matrix of a ($k$, $n$)-linear block code, and acts as an encoder. In this way, each message is mapped to a unique codeword. 

Scrambling is proposed in \cite{nonsys} for PHY security in which Alice implements encoding as $
    \mathbf{c}=\mathbf{u}\mathbf{S}\mathbf{G}
$,
where $\mathbf{G}$ is the $k\times n$ generator matrix in systematic form and $\mathbf{S}$ is a
nonsingular $k\times k$ binary scrambling matrix. 
Owing to its systematic form, $\mathbf{G}$ can be written
as $\mathbf{G}=[\mathbf{I}|\mathbf{C}]$, where $\mathbf{I}$ is a $k\times k$ identity matrix and $\mathbf{C}$ is $k\times (n-k)$ representing
the parity-check constraints. 
Thus, encoding simply consists of replacing the information vector
with its scrambled version $\mathbf{u}'= \mathbf{u}\mathbf{S}$, and then applying the linear block code given by $\mathbf{G}$.
According to the physical layer security principle, both $\mathbf{S}$ and $\mathbf{G}$ are made public and both of
them are necessary for decoding. 
Decoding process includes descrambling, i.e., multiplication by $\mathbf{S}^{-1}$, which will propagate the errors \cite{ScramblingBaldi} at lower SNRs where Eve is assumed to be working while leaving the decoding performance at higher SNRs (Bob's intended region) ``less affected".

Randomized encoding scheme, also known as coset coding, is the classical coding method for confusing Eve. Unlike two previous methods, where each codeword corresponded to a unique message, the randomized scheme maps each message to a coset of codewords. 
The randomized encoding is devised in the following manner. Let $\mathbf{m}$ and $\mathbf{r}$ denote vectors of message and random bits of length $k$ and $r$, respectively. Furthermore, consider two matrices  $\mathbf{G}$ and $\mathbf{H}$ of size $k\times n$ and $r\times n$, respectively. Then, a codeword of length $n$ is generated through $
   \mathbf{c}= \begin{bmatrix} \mathbf{m}&\mathbf{r} \end{bmatrix}\begin{bmatrix} \mathbf{H}\\\mathbf{G} \end{bmatrix},
$
assuming the rows of $\mathbf{G}$ and $\mathbf{H}$ are linearly independent. Denoting the coset corresponding to the $i$th message, i.e., $\mathbf{m}^i$, by $\mathcal{C}^i$, the randomized encoder picks a codeword from $\mathcal{C}^i$ based on $\mathbf{r}$.
For all these schemes, no secret information (e.g., keys) is assumed between Alice and Bob, and Bob and Eve are both aware of the underlying coding technique employed by Alice.
\section{Deep Learning Tools for Code Design}\label{sec:Deep learning tools for code design}
We present an overview of two deep learning tools that are essential for designing secure codes. The first one is a framework for the computation of mutual information between two random variables with arbitrary probability distributions. The second one is a technique that enables training of a neural network with binary constraints imposed on its outputs/weights.
\subsection{Mutual Information Neural Estimation (MINE)}\label{Sec:Mine: Mutual Information Neural Estimation}
As described in Section \ref{sec:Channel Model}, one needs to compute mutual information in order to measure the security performance. For continuous channels including AWGN, computation of mutual information is not tractable in general \cite{Deepshit}. However, a recently-proposed algorithm in deep learning literature, i.e., MINE, enables us to compute this quantity using SGD and Donsker-Varadhan representation of the Kullback Leibler (KL) divergence defined as \begin{align} \nonumber \label{DV}
    \mathbb{I}(\textbf{x};\textbf{y}) &= \mathbb{D}_{\text{KL}}(p(\textbf{x},\textbf{y})||p(\textbf{x})p(\textbf{y}))\\
                   & \geq \mathbb{E}_{p(\textbf{x},\textbf{y})} [U] - \log(\mathbb{E}_{p(\textbf{x})p(\textbf{y})}[e^{U}])
\end{align}
where $\textbf{x}$ and $\textbf{y}$ denote two random variables with arbitrary distributions. Furthermore, $U$ is a function that maps the samples from the joint and marginal distributions to a real number. Given that $U$ is expressive enough, the above lower bound converges to the true mutual information. Denoting the parameterized version of $U$ by $\{U_{\mathbf{\gamma}}\}_{\mathbf{\gamma} \in {\Gamma}}$, MINE solves the following optimization problem:
\begin{equation}
\label{eq:MINE}
\underset{\mathbf{\gamma} \in \Gamma}{\text{sup}}~~ \mathbb{E}_{p(\textbf{x},\textbf{y})} [U_{\gamma}] - \log(\mathbb{E}_{p(\textbf{x})p(\textbf{y})}[e^{U_{\gamma}}]),
\end{equation}
where the expectations are estimated using empirical samples from the joint and marginal distributions. This can be effectively solved via SGD algorithms using mini-batches of two datasets corresponding to each distribution. In this way, the neural network is trained through back-propagation which results in the optimal set of parameters $\mathbf{\theta}$. 
If one chooses $U_{\gamma}$ from the class of functions represented by fully-connected feed-forward neural networks, the size of the function space $\Gamma$ is determined by the number of hidden layers and neurons in each network. If this numbers are chosen properly, $\Gamma$ is large enough for an accurate estimate of the mutual information. In practice, it is important to collect enough data from each distribution for training in order to ensure convergence \cite{MINE}.

\subsection{Neural Networks with Binarized Outputs}\label{Sec:Neural networks with binarized outputs}
As we will show in section \ref{Sec:Secure Code Design via Deep learning}, learning end-to-end secure codes for the wiretap channel requires utilizing neural networks which output binary values $0$, $1$ (or equivalently $+1$, $-1$). This can be realized by using a sign function 
$
    x^b=\begin{cases}
    1\ ,x\geq 0,\\
    0\ , x<0,
    \end{cases}
$
as the activation function of the output layer. However, derivative of the sign function is zero almost everywhere, making it apparently incompatible with the backpropagation, as the exact gradient of the cost with respect to the neural network's weights/biases would be zero. To get around this impediment, the authors in \cite{binaryNN} propose a method called straight-through estimator (STE) which affects the training process in two phases. Firstly for the forward pass, the binary value $x^b$ is evaluated using the sign function. Secondly for the backpropagation, the gradient of the loss function with respect to (w.r.t) $x^b$ is considered to be the same as that w.r.t
$x$, i.e., the gradient of the sign function is ignored. This heuristic is shown to work well in various learning settings \cite{binaryNN}. 

\section{Secure Code Design via Deep Learning }\label{Sec:Secure Code Design via Deep learning}
In this section, we study how deep learning can be employed for designing codes that satisfy the reliability and security constraints described in Section \ref{sec:Channel Model}. To this end, we consider two approaches. The first one is based on the idea of autoencoders and is presented recently in \cite{autoencoderPHY}. The second one is our proposed method which is called end-to-end learning. Although both these approaches employ DNNs for learning secure codes, the approach in \cite{autoencoderPHY} uses a different metric for measuring security and the structure of the employed DNNs are also different. Furthermore, in order to comply with the finite alphabet signaling for achieving  physical layer security \cite{OurSurvey}, we force the DNNs to learn codes which are being modulated via BPSK scheme (either $+1$ or $-1$ are being sent through the channel). As will be discussed later, this is in contrast to the approach presented in \cite{autoencoderPHY} which assumes a continuous modulation is being used at the encoder and real-valued symbols are being sent through the channel. 
\subsection{Secure Code Design via Autoencoders}
The authors in \cite{autoencoderPHY} propose learning codes for the Gaussian wiretap channel utilizing the autoencoders. These networks can be seen, in general, as two DNNs where the input of the first network and the output of the second network are identical. An autoencoder matches well to the classic problem of communication between two parties where its first constituent DNN can be seen as the encoder and the second one corresponds to the decoder. Denoting the secret message bits and the Eve's observation by $\mathbf{m}$ and $\mathbf{z}$, respectively, the optimization problem which is being solved by the autoencoder in \cite{autoencoderPHY} is
\begin{equation}
    \min_C\ w_B \text{BER}+ w_E \mathbb{I}(\mathbf{m},\mathbf{z}),
\end{equation}
where the BER corresponds to the reliability between Alice and Bob, while $w_B$ and $w_E$ are weights which indicate the relative importance of each term in the objective function. The encoder network of the autoencoder learns the modulated version of the codes learns based on the above objective function. This modulation does not correspond to a finite-alphabet scheme as the output of the encoder consists of the real-valued numbers. The authors in \cite{autoencoderPHY} estimate BER for each mini-batch of data via mean square error. For the computation of mutual information term, two different approximations are being used by the authors. The first method uses an upper bound of the leakage based on the upper and lower bounds of differential entropy of Gaussian mixtures, while the second one approximates the leakage function by invoking the Taylor expansion. There are issues associate with each of the two methods for computing the mutual information. Specifically, the upper bound approach is shown to be not very effective during training as it is loose. Moreover, one needs to compute the entropy of the Gaussian mixtures for the Taylor expansion scheme which is not trivial in general. Therefore, the authors had to force their analysis to the low-SNR regimes where the Gaussian mixtures can be approximated by a single Gaussian and computation of the entropy becomes feasible. We also note that the authors in \cite{autoencoderPHY} do not consider any learning capability (in terms of DNNs) for Eve and assume that learning performance between Alice and Bob would not be affected by the presence of an Eve which could also train a DNN to decode the messages simultaneously. This is an important note as the adversary may have access to excessive computational resources which could deteriorate the learning performance of the autoencoder in finding secure codes.

\subsection{Learning End-to-End Secure Codes}
\label{sec:Encoder's architectures}
\begin{figure}
\centering
\includegraphics[scale=0.3]{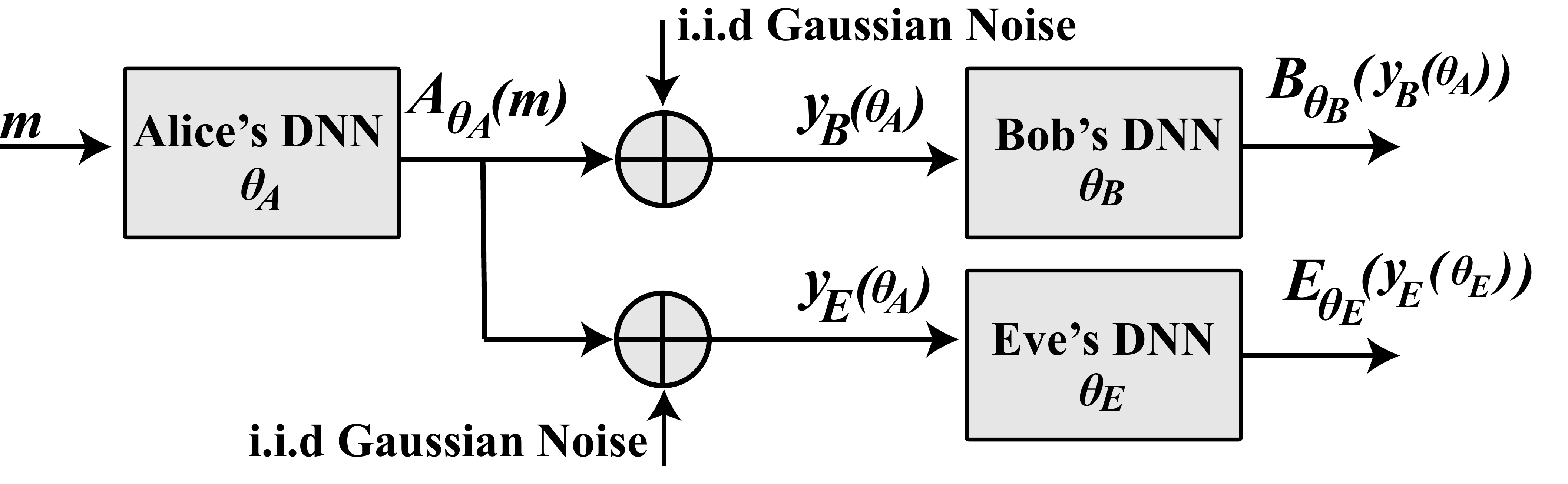}
\caption{End-to-end learning of the encoder and the decoders in the Gaussian wiretap channel.
}
\label{fig:Model}
\end{figure}

We assume DNNs are being used at each of the communication parties in the wiretap channel, i.e., Alice, Bob and Eve, as illustrated in Fig. \ref{fig:Model}. Specifically, DNNs at Alice encode/map a message $\mathbf{m}$ to a codeword $\mathbf{c}$, corresponding noisy versions of which are being decoded by the DNNs at Bob and Eve. In order to ensure that DNN at Alice sends BPSK symbols through the channel, we use STE described in Section \ref{Sec:Neural networks with binarized outputs} in its output layer.

As communicating parties are modeled by DNNs, it is crucial for the learning process to have a proper loss function based on the design objectives. The main difference of our approach to the autoencoder approach in \cite{autoencoderPHY} relies on the fact that we make use of MINE to directly compute the mutual information through the gradient decent algorithm. Towards this goal, we begin with introducing notations. We denote the output of the encoder at Alice by $A_{\mathbf{\theta}_A}(\mathbf m)$ where $\mathbf m$ and $\mathbf{\theta}_A$ denote the input bits (secret message) and DNNs' parameters at Alice, respectively. We note that $A_{\mathbf{\theta}_A}(\mathbf m)$ is a vector of $+1$'s and $-1$'s, corresponding to the BPSK version of the learned codeword by Alice, which is generated utilizing the STE method described in Section \ref{Sec:Neural networks with binarized outputs}. Similarly, the outputs of the decoders at Bob and Eve are shown by $B_{\mathbf{\theta}_B}(\mathbf{y}_B(\theta_A))$ and $E_{\mathbf{\theta}_E}(\mathbf{y}_E(\theta_A))$ where $\mathbf{y}_B(\theta_A)$ and $\mathbf{y}_E(\theta_A)$ denote the observations at Bob and Eve, respectively. For Alice-Bob link, we wish to design an encoder at Alice which satisfies the reliability and security constraints. In particular, the reliability constraint requires the DNN at Bob to be able to decode/map $\mathbf{y}_B(\theta_A)$ to the secret message $\mathbf m$. Therefore, the reliability loss function is defined as
\begin{equation}
\label{eq:ReliabilityLoss}
    L_R(\theta_A,\theta_B)=d\big(\mathbf{m},B_{\mathbf{\theta}_B}\big(\mathbf{y}_B(\theta_A)\big)\big),
\end{equation}
where $d$ denotes the $L1$ distance function which is defined for vectors $\mathbf{u}$ and $\mathbf{v}$ of length $k$ as $d(\mathbf{u},\mathbf{v})=\sum_{i=1}^k |\mathbf{u}_i-\mathbf{v}_i|$ assuming $\mathbf{u}_i$ to be the $i$th element of $\mathbf{u}$. The security constraint is measured in terms of the mutual information between the Eve's observation and the messages. Therefore, \begin{equation}
\label{eq:SecurityLoss}
     L_S(\theta_A)=\mathbb{I}_{\phi}\big(\mathbf{m};\mathbf{y}_E(\theta_A)\big),
\end{equation}
is defined as the security loss function, where $\mathbb{I}_{\phi}$ is computed using MINE described in Section \ref{Sec:Mine: Mutual Information Neural Estimation} via a fully-connected feed-forward DNN parameterized by $\phi$. Furthermore, all the message bits are expected to be encoded in $A_{\mathbf{\theta}_A}(\mathbf m)$, which leads to Alice's loss function defined as
\begin{equation}
\label{eq:AliceLoss}
     L_A(\theta_A)=-\mathbb{I}_{\psi}\big(\mathbf{m};A_{\mathbf{\theta}_A}(\mathbf{m})\big),
\end{equation}
where $\mathbb{I}_{\psi}$ denotes the mutual information estimation using MINE via the same type of DNN as for $\mathbb{I}_{\phi}$, which is denoted by $\psi$ for this case. The loss function for the Alice-Bob link is composed of the above three losses as
\begin{equation}
\label{eq:Alice-BobLoss}
    L_{AB}(\theta_A,\theta_B)=\alpha L_R(\theta_A,\theta_B)+\beta L_S(\theta_A)+\gamma L_A(\theta_A),
\end{equation}
where the parameters $\alpha$, $\beta$ and $\gamma$ reflect the relative gains for each constituent loss. Then, the optimal DNNs at Alice and Bob, denoted by $\theta^*_A$, $\theta^*_B$, respectively, are obtained by
\begin{equation}
    (\theta^*_A, \theta^*_B)=
    \argmin_{\theta_A,\theta_B}\  L_{AB}(\theta_A,\theta_B).
\end{equation}
Meanwhile, Eve aims at decoding the message bits by choosing the loss function,
\begin{equation}
\label{eq:EveLoss}
    L_E(\theta_A,\theta_E)=d\big(\mathbf{m},E_{\mathbf{\theta}_E}\big(\mathbf{y}_E(\theta_A)\big)\big).
\end{equation}
Similarly, optimal Eve is obtained by $ \theta^*_E=
    \argmin_{\theta_E}\  L_{E}(\theta_A,\theta_E).$
For the DNNs at Bob and Eve, denoted by $\theta_B$ and $\theta_E$, respectively, we use fully-connected feed-forward networks details of which are presented in Section \ref{sec:Numerical Results}. We have not observed performance improvements upon utilizing a different DNN structure for these decoders. In contrast, we study three different DNN structures for Alice. On the other hand, choice of the encoder's DNNs have been shown to play a major role on the overall performance of the system as pointed out in the following.

We consider different architectures for the DNN at Alice based on each of the coding methods described in Section \ref{sec:Coding schemes for securing the PHY layer}, and compare their performance in Section \ref{sec:Numerical Results}.
\subsubsection{Classic encoder}
\label{sec:Classic encoder}
Alice maps the input bits to the output bits via a DNN consisting of fully-connected layers followed by convolutional layers as depicted in Fig. \ref{fig:Enocders}.a. The reshape operation refers to transforming a vector of size $n$ to an $i\times j$ matrix ($n=i\times j$), which allows output of the fully-connected layers to be further processed by the convolutional layers. The reverse of such operation is referred to as flattening. We have observed through several numerical experiments that this concatenation results in the best performance for designing secure PHY codes. The reverse order, i.e., putting fully-connected layers after the convolutional ones, is widely used for the case of image classification. 
\subsubsection{Scrambling encoder}
\label{sec:Scrambling encoder}
\begin{figure}
\centering
\includegraphics[scale=0.33]{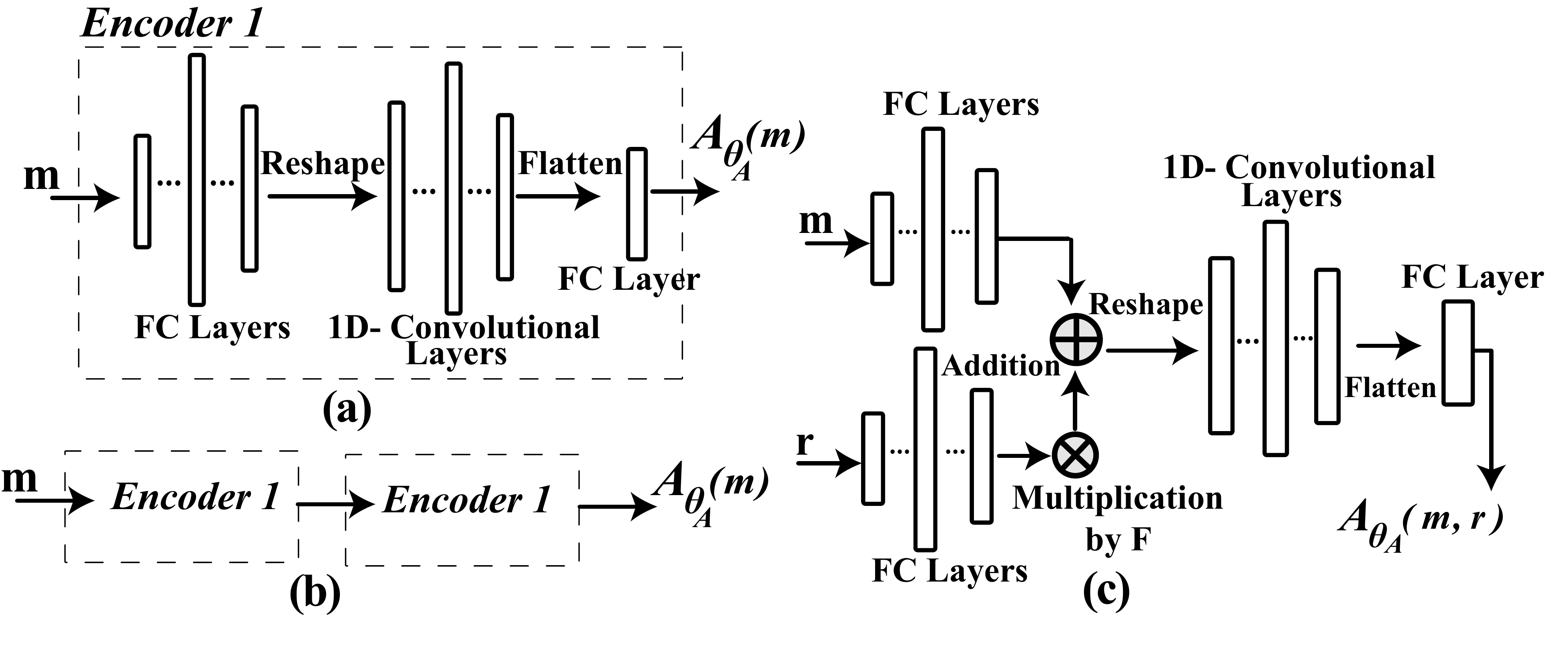}
\caption{Three different DNN architectures for the encoder at Alice (FC: fully-connected).}
\label{fig:Enocders}
\end{figure}

In this case, as illustrated in Fig. \ref{fig:Enocders}.b the codewords generated by the classic encoder are input to another DNN (of the same architecture) which acts as a scrambler, and generates the final codeword sent over the channel. Although the generator matrix in the scrambling method, described in Section \ref{sec:Coding schemes for securing the PHY layer}, is systematic, we refrain from imposing further constraints on the DNNs as we have not seen any performance improvements by doing so. In fact, we let the DNNs find the optimal encoder and the optimal scrambler which would minimize the loss function.

\subsubsection{Randomized encoder}
\label{sec:Randomized encoder}
This encoder is constructed based on the idea of coset coding where the random bits ($\mathbf{r}$) are involved in the process of encoding the message bits ($\mathbf{m}$). Fig. \ref{fig:Enocders}.c demonstrates this encoder,  where separate DNNs are used for the random and the message bits, and the weighted sum of the outputs are input to the convolutional layers. Specifically, the encoder is now denoted by $A_{\mathbf{\theta}_A}(\mathbf{m},\mathbf{r})$. We note that the coset coding scheme as described in Section \ref{sec:Coding schemes for securing the PHY layer} involves mod-$2$ sum of the codewords generated from the random and the message bits, which enables constructing the coset structure. However, this sum cannot be used for learning codes in an end-to-end fashion as the mod-$2$ sum does not allow the gradient flow as required by the back-propagation algorithm. Hence, we resort to the weighted sum as a workaround, to mix the DNNs' outputs. Furthermore, in order to encourage the incorporation of the random bits during the encoding process, we modify the reliability loss function in (\ref{eq:ReliabilityLoss}) as 
\begin{align}
    \label{eq:ReliabilityLossRandom}
     L_R(\theta_A,\theta_{B,\mathbf r},\theta_{B,\mathbf m})=&d\big(\mathbf{m},B_{\mathbf{\theta}_{B,\mathbf m}}\big(\mathbf{y}_B(\theta_A)\big)\big)\nonumber\\&+d\big(\mathbf{r},B_{\mathbf{\theta}_{B,\mathbf r}}\big(\mathbf{y}_B(\theta_A)\big)\big),
\end{align}
where $B_{\mathbf{\theta}_{B,\mathbf m}}$ and $B_{\mathbf{\theta}_{B,\mathbf r}}$ denote the decoders corresponding to the message and random bits, parameterized by $\theta_{B,\mathbf m}$ and $\theta_{B,\mathbf r}$, respectively. So, Bob is trained to decode both the message and random bits from the observation $\mathbf{y}_B$. The following algorithm presents the details of the learning process for this encoder.

\begin{algorithm}[t!]
\caption{End-to-end learning with the randomized encoder}
$\theta\gets$ Initialization values, for $\theta\in\{\theta_A,\theta_{B,\mathbf m},\theta_{B,\mathbf r},\theta_E,\psi,\phi\}$
\begin{algorithmic}
\Repeat
\State Generate $b$ samples for the message and the random bits:
\State$\{\mathbf{m}^{(i)}\}_{i=1}^b$, $\{\mathbf{r}^{(i)}\}_{i=1}^b$
\State Generate $\mathbf{y}_B^{(i)}$ and $\mathbf{y}_E^{(i)}$ via $\mathbf{c}^{(i)}=A_{\mathbf{\theta}_A}(\mathbf{m}^{(i)},\mathbf{r}^{(i)})$.
\State $\{\Tilde{\mathbf{m}}^{(i)}\}_{i=1}^b\gets$ Shuffled $\{\mathbf{m}^{(i)}\}_{i=1}^b$ w.r.t $i$'s
\State Generate $B_{\mathbf{\theta}_{B,\mathbf m}}(\mathbf{y}_B^{(i)})$, $B_{\mathbf{\theta}_{B,\mathbf r}}(\mathbf{y}_B^{(i)})$, $E_{\mathbf{\theta}_{E}}(\mathbf{y}_E^{(i)})$
\State Compute $L_{AB}^{(i)}(\theta_A,\theta_B)$ via (\ref{eq:Alice-BobLoss}), $\theta_B=\{\theta_{B,\mathbf m},\theta_{B,\mathbf r}\}$
\State Compute $L_E^{(i)}(\theta_A,\theta_E)$ via (\ref{eq:EveLoss})
\State $\mathcal{V}(\theta_A,\theta_B)\gets \frac{1}{b}\sum_{i=1}^b L_{AB}^{(i)}(\theta_A,\theta_B)$
\State The gradient $\mathcal{G}(\theta_A,\theta_B)\gets\nabla_{(\theta_A,\theta_B)}\mathcal{V}$
\State $\mathcal{U}(\theta_A,\theta_E)\gets \frac{1}{b}\sum_{i=1}^b L_E^{(i)}(\theta_A,\theta_E)$, $\mathcal{G}(\theta_E)\gets\nabla_{\theta_E\ }\mathcal{U}$

\State Update the networks' parameters:
\State $\theta\gets \theta-\mathcal{G}(\theta)$ for $\theta\in\{\theta_A,\theta_B,\theta_E\}$\\

    $\ \ \ \xi(\psi)\gets\frac{1}{b}\sum_{i=1}^{b}T_\psi(\mathbf{c}^{(i)},\mathbf{m}^{(i)})\text{-}\log (\frac{1}{b}\sum_{i=1}^b e^{T_\psi(\mathbf{c}^{(i)},\Tilde{\mathbf{m}}^{(i)})})$\\
$\ \ \ \vartheta(\phi)\gets\frac{1}{b}\sum_{i=1}^{b}T_\phi(\mathbf{y}_E^{(i)},\mathbf{m}^{(i)})\text{-}\log (\frac{1}{b}\sum_{i=1}^b e^{T_\phi(\mathbf{y}_E^{(i)},\Tilde{\mathbf{m}}^{(i)})})$
\State MINE gradients: $\mathcal{G}(\psi)\gets\nabla_{\psi\ }\xi(\psi)$, $\mathcal{G}(\phi)\gets\nabla_{\phi\ }\vartheta(\phi)$
\State $\theta\gets \theta+\mathcal{G}(\theta)$ for $\theta\in\{\psi,\phi\}$
\Until{Convergence}
\end{algorithmic}
\label{Alg:Algorithm1}
\end{algorithm}

\subsection{Adaptive Gradient Clipping}
According to the loss function in (\ref{eq:Alice-BobLoss}), the encoder's parameters, $\theta_A$, are updated by three gradients, two of which correspond to the use of MINE. We have seen through our experiments that these gradients can overwhelm the other one. This indeed can lead to a point where the encoder puts all its attention on the mutual information losses \cite{MINE}, and ignore the reliability loss. To circumvent this impediment, we adaptively clip the MINE-originated gradients, $\mathcal{G}_M$, during the learning process so that their $L2$-norm is at most equal to that of the gradient of the
reliability loss, $\mathcal{G}_R$. Mathematically, for the adapted $\mathcal{G}_M$, denoted by $\mathcal{G}_a$, we have
\begin{equation}
    \mathcal{G}_a=\min(||\mathcal{G}_R||_2,||\mathcal{G}_M||_2)\frac{\mathcal{G}_M}{||\mathcal{G}_M||_2}.
\end{equation}
\section{Numerical Results}\label{sec:Numerical Results}
In this section, we present numerical results on the performance of the codes learned in an end-to-end manner utilizing deep learning. We begin by presenting the specifications of the DNNs described in  Section \ref{Sec:Secure Code Design via Deep learning} which are observed to be efficient in terms of complexity and performance based on several experiments. The DNNs for MINE which were denoted by $\psi$ and $\phi$, consist of $4$ fully-connected hidden layers each having $400$ neurons and rectified linear unit (ReLU) as the activation function. For the Encoder $1$ structure in Fig. \ref{fig:Enocders}, we use $3$ hidden layers with $500$ neurons and ReLU activation function, followed by another layer with $256$ neurons and tanh function. Furthermore, we use two $1$D convolutional layers with ReLU activation function whose kernel size (length of the convolution window) is $4$ with a stride of $1$. The number of output filters for the first and the second one is set to $16$ and $32$, respectively. Dimension of the very last fully-connected layer after the convolutional layers is the same as the code length ($n$) for which the STE is used to binarized the outputs. For the randomized encoder in Fig. \ref{fig:Enocders}.c, there are two fully-connected networks and one convolution network whose specifications are set to be the same as the corresponding ones in Encoder $1$. The factor $F$ is set to $20$ for this encoder. Finally, the DNNs for the decoders, denoted by $\theta_B$ and $\theta_E$, are both chosen to have $5$ hidden layers each having $500$ neurons and ReLU as the activation function. For their output layer, the sigmoid function is used where the values greater than $0.5$ are decoded as $1$, and those less than $0.5$ are mapped to $0$. We have not seen any performance improvement for Eve (in terms of decoding) by choosing a more complicated network as $\theta_E$. For training, we initialize the DNN parameters with random values following a Gaussian distribution with zero mean and variance $0.1$. We further use a minibatch of size $b=2048$ along with Adam optimizer with a learning rate of $0.0004$ to minimize the loss function. We note that the training alternates between Alice-Bob and Eve, each being trained on one minibatch, as described in detail in Algorithm \ref{Alg:Algorithm1} for the randomized encoder. A Tensorflow implementation of Algorithm \ref{Alg:Algorithm1} is presented as a Github repository in \cite{mycode}. 

\begin{figure}

\centering
\includegraphics[width=11cm]{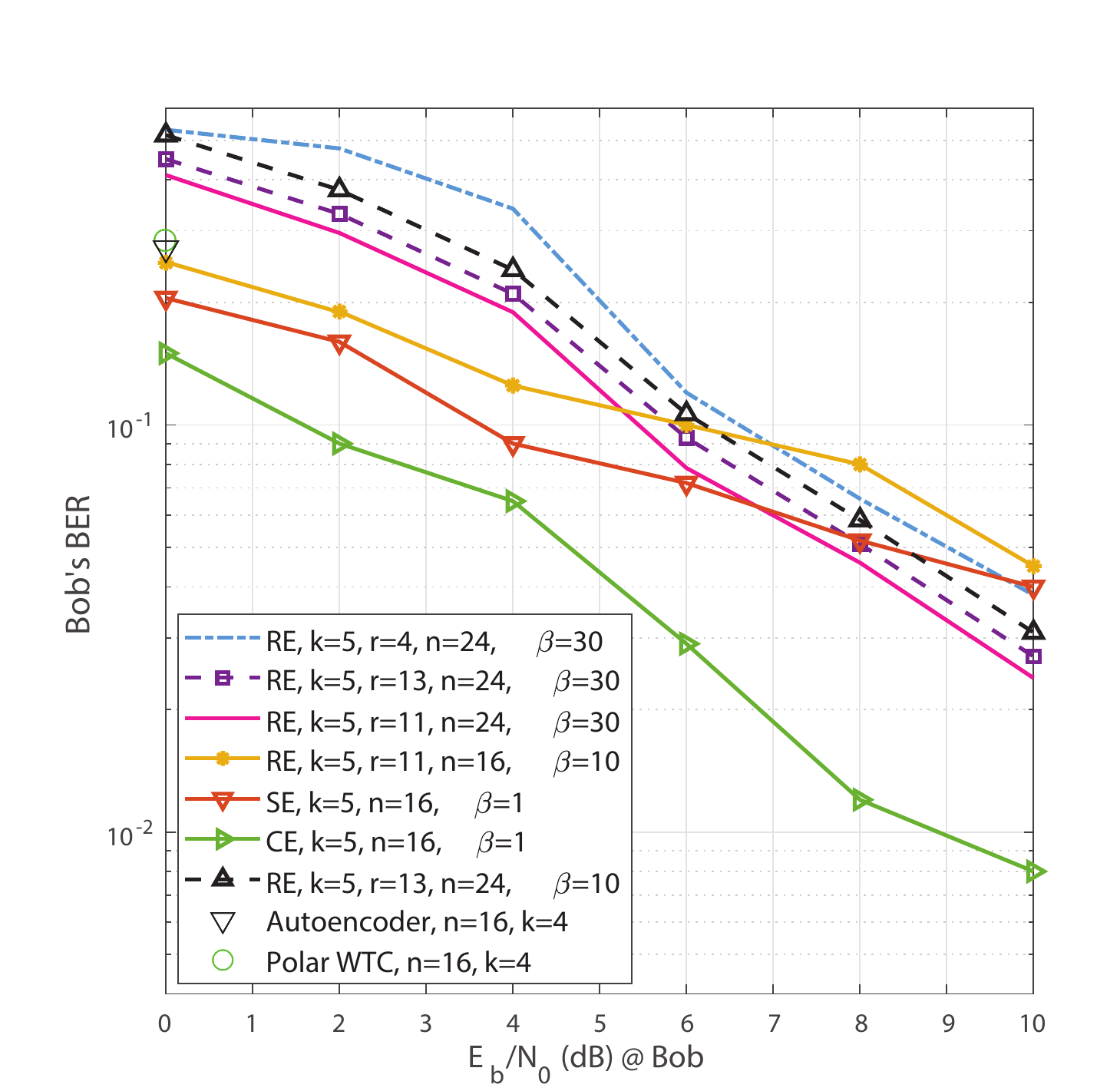}
\caption{BER performance of the end-to-end learned codes at Bob for the case where $\alpha=\gamma=1$. CE, SE and RE denote classical, scrambling and randomized encoder, respectively.}
\label{fig:BERs}

\end{figure}

\begin{figure}

\centering
\includegraphics[width=11cm]{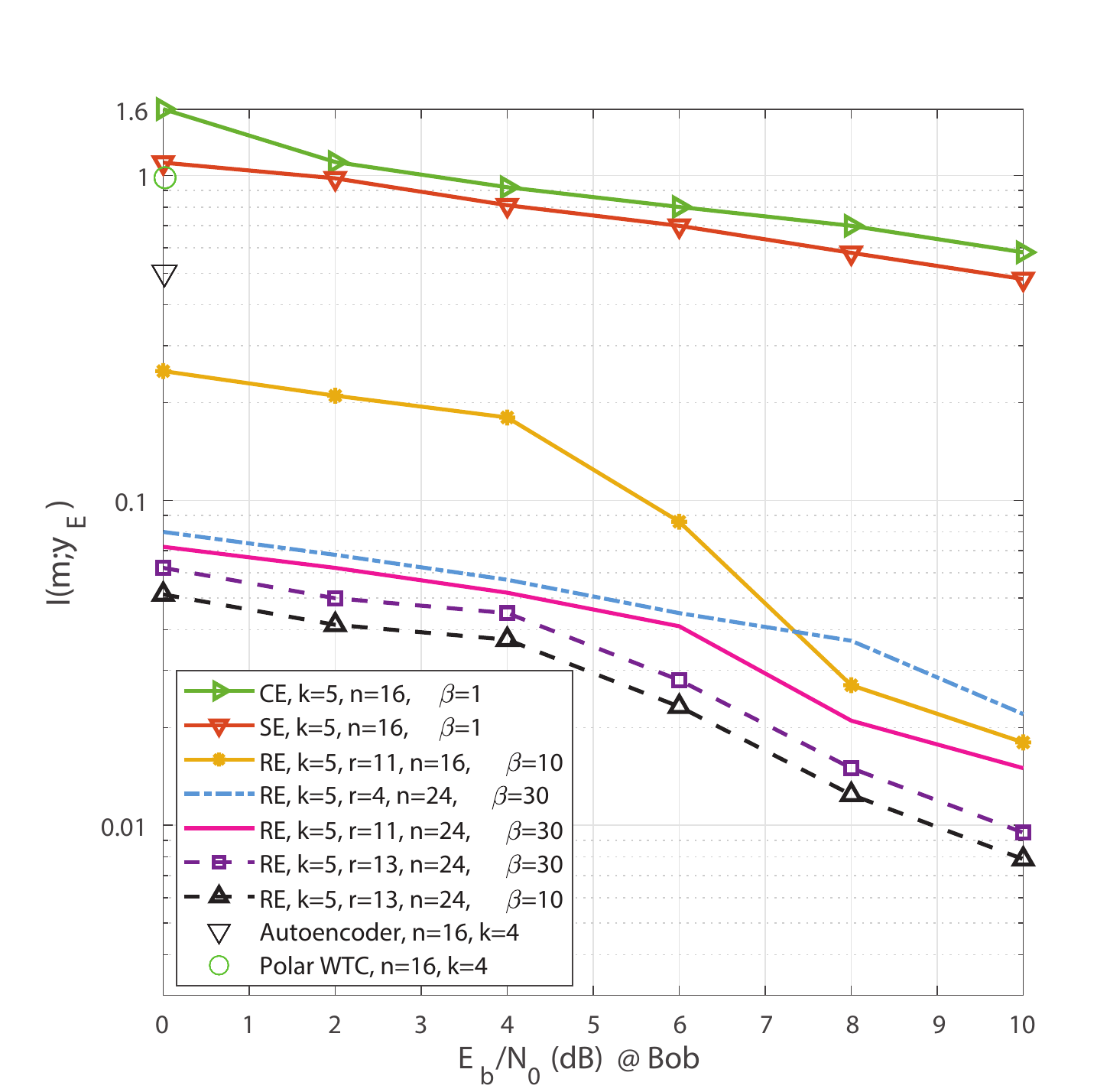}
\caption{Mutual information between the Eve's observation and the messages for the end-to-end learned codes.}
\label{fig:MIs}
\end{figure}

We have examined the performance of the three encoders presented in Section \ref{sec:Encoder's architectures} along with the effect of different code parameters, i.e., $n$, $k$ and $r$. Upon increasing the dimension of the involved random variables, MINE requires a larger sample space and a more complicated network to converge to the right value which could be prohibitively complex. Due to this fact, we mainly focus on small length codes and evaluate their performances from the perspective of security and reliability constraints. This is demonstrated in Figs. \ref{fig:BERs} and \ref{fig:MIs} for BER at bob (reliability) and $\mathbb{I}(\mathbf m;\mathbf{y}_E)$ (security), respectively, when Eve's SNR is set to $-2$ dB. We note that training is done for Alice-Bob and Eve at a specific SNR at Bob which is represented in the x-axis. Furthermore, the weights in the loss (\ref{eq:Alice-BobLoss}), i.e., $\alpha$, $\beta$ and $\gamma$, are chosen in a way to ensure that all the $k$ secret bits are used for encoding, i.e., $\mathbb{I}(\mathbf{m},A_{\theta_A}(\mathbf{m}))=k$. 

A code designer considers Figs. \ref{fig:BERs} and \ref{fig:MIs} simultaneously, and wishes to select the codes which satisfy the predefined requirements on the reliability and security constraints. Specifically, for $k=5$, $n=16$ and $E_b/N_0=10$ dB, the classic encoder (CE), described in Section \ref{sec:Classic encoder}, has been capable of learning codes which result in an approximate BER of $0.008$. However, it performs poorly from the security perspective as $\mathbb{I}(\mathbf m;\mathbf{y}_E)$ is around $0.7$, and Eve's BER is $0.26$. At the expense of sacrificing the reliability, the scrambler encoder (SE) introduced in Section \ref{sec:Scrambling encoder} has shown to improve the security loss by a small margin for the same design parameters. The poor performance of the CE and SE in providing secure communication stems from their DNNs' architecture which prevents them from learning secure codes. This point is further highlighted by looking at the performance of the randomized encoder (RE) proposed in Section \ref{sec:Randomized encoder}. One can see that for $k=5$, $r=11$, $n=16$ and $E_b/N_0=10$ dB, $\mathbb{I}(\mathbf m;\mathbf{y}_E)$ of about $0.02$ is achieved which exhibits a great improvement in comparison to the other two encoders. For this case, we have also observed that BER at Eve is above $0.49$ which indicates that she has not been able to learn a decoder for the messages. Of course, the gain in security is achieved by compromising the reliability constraint to some extent which might be tolerated based on the design goals. 

Figs. \ref{fig:BERs} and \ref{fig:MIs} also include the performance results from the autoencoder approach, presented in \cite{autoencoderPHY}, as the most similar learning approach for physical layer code design to our work and the polar wiretap coding (WTC) \cite{polar} as a state-of-the-art reference point. For these schemes, Eve is assumed to work at $\text{SNR} = -5$ dB while Bob's SNR is $0$ dB. As noted in \cite{autoencoderPHY}, the authors are limiting their analysis to the low SNR regime as this assumption enables them to approximate the mutual information term. It is shown that for similar code lengths and code rates, the codes learned through the end-to-end manner via the randomized encoder proposed in Section \ref{sec:Randomized encoder} outperform both the codes corresponding to the autoencoder and the polar WTC schemes from the security and reliability perspectives.

Another important observation from the results can be made in relation to the $r$, i.e., the number of random bits. Specifically, for the RE with $k=5$, $n=24$, it is shown that as $r$ gets larger, the system gets more secured. Notably, $\mathbb{I}(\mathbf m;\mathbf{y}_E)$ less than $0.01$ is achieved via $k=5$, $r=13$, $n=24$ while BER at Bob is less than $0.03$. To get a better sense of the security performance of the learned codes, we consider the equivocation region which is defined as $\mathbb{H}(\mathbf m|\mathbf{y}_E)/k$ versus the transmission rate $k/n$. For the rate of $5/24$, and when the SNRs at Eve and Bob are set to $-2$ and $10$ dB, respectively, the boundary of this region is the maximum value, i.e., $1$. One can confirm that the points achieved by the randomized encoders with parameters $r=11$ and $r=13$ are $0.995$ and $0.997$, respectively, which are very close to the boundary. Finally, we note that a designer can tweak the weights $\alpha$, $\beta$ and $\gamma$, in order to learn codes which are suited for the desired levels of reliability and security. For instance, it is shown that increasing $\beta$ leads to a lower $\mathbb{I}(\mathbf m;\mathbf{y}_E)$ and a higher BER at Bob. 

\section{Conclusions}
\label{sec:Conclusions}
We have proposed end-to-end learning of the codes for the Gaussian wiretap channel where each party is equipped with DNNs. Alice and Bob are trained in a manner to provide reliability for Bob while ensuring security against Eve. Simultaneously, Eve trains her network in order to decode the secret messages. The security is measured in terms of the mutual information between the messages and the Eve's observation using MINE. The numerical results show that a secure system can be designed in this setting with finite-length codes. In particular, by utilizing a randomized encoder, Alice can achieve points as close as $0.003$ to the boundary of the equivocation region for a transmission rate of $5/24$. The presented results are compared with the existing deep learning approach based on the idea of autoencoders which highlights the superiority of our proposed method in achieving reliable and secure codes for the wiretap channel.
\newpage
\bibliographystyle{IEEEtran}  
\bibliography{main-reject}
\end{document}